\documentclass[conference]{IEEEtran}
\IEEEoverridecommandlockouts
\usepackage{cite}
\usepackage{amsmath,amssymb,amsfonts}
\usepackage{algorithmic}
\usepackage{algorithm}
\usepackage{graphicx}
\usepackage{textcomp}
\usepackage{xcolor}
\usepackage{enumitem}
\usepackage{booktabs}
\usepackage{tabularx}
\usepackage{pifont}
\usepackage{balance}
\usepackage{tablefootnote}
\usepackage{multicol}

\def\BibTeX{{\rm B\kern-.05em{\sc i\kern-.025em b}\kern-.08em
    T\kern-.1667em\lower.7ex\hbox{E}\kern-.125emX}}
\begin{document}

\title{Self-Evolving Beyond Human-Crafted Experience for EDA Scripting}


\author{
\IEEEauthorblockN{Yang Liu$^{1,*,\dagger}$, Shengbo Tong$^{1,2,*}$, Xiyuan Chen$^{1,3}$, Shiwei Hou$^{1}$, Sen Yuan$^{1}$,}
\IEEEauthorblockN{Hao Wang$^{1}$, Yanfeng Li$^{1,\dagger}$, Hongwen Li$^{1,\dagger}$, Yiming Zhu$^{1}$, Zhuolun He$^{2,\dagger}$}, Wenjian Yu$^{2,\dagger}$
\IEEEauthorblockA{
$^{1}$CXMT, $^{2}$Tsinghua University, $^{3}$USTC\\
paul.liu@cxmt.com, tsb23@mails.tsinghua.edu.cn\\
$^{*}$These authors contributed equally to this work.\\
$\dagger$Corresponding authors.
}
}

\maketitle

\begin{abstract}
The growing complexity of electronic systems demands efficient design flows, yet EDA scripting remains a bottleneck: proprietary APIs are sparsely documented, and errors often appear as silent failures or cryptic exceptions. Hand-crafted knowledge helps but is difficult to maintain as toolchains evolve. We present \textbf{ZhuLong-SE (Self-Evolution)}, an execution-grounded LLM agent that learns reusable knowledge from prior task traces. Building on ZhuLong~\cite{liu2026zhulong}, it enhances the task-level harness with API retrieval, documentation responses, isolated sandbox execution, and static checking, then adds governed cross-task self-evolution.
Beyond solving individual tasks, ZhuLong-SE distills past execution traces into a three-tier memory store governed by ground-truth anchoring, disconfirmation filtering, held-out rollback, and anti-leakage retrieval. On the 158-task EDA-Eval-PyAether benchmark, the full task-level harness, with self-evolution disabled, achieves 92.4\% pass@1 and outperforms ZhuLong on all five tested LLM backbones. On a harder instance-disjoint test set, self-evolution from the bare configuration surpasses a human-crafted-experience baseline by 2.5 percentage points (80.0\% vs.\ 77.5\% pass@1).
\end{abstract}

\begin{IEEEkeywords}
System-level design methodologies, design knowledge reuse, LLM agents, EDA scripting, continual learning, self-evolution, memory reliability, PyAether.
\end{IEEEkeywords}

\section{Introduction}
\label{sec:introduction}

EDA scripting is challenging because tool APIs are tool-specific, sparsely documented, and exposed through SWIG bindings whose type rules differ from native Python. ZhuLong~\cite{liu2026zhulong} is the first work that established an execution-grounded LLM agent for EDA scripting that combines API retrieval, documentation inspection, and sandbox execution via unified MCP tools, augmented by an offline API self-exploration mechanism that infers undocumented API behaviors through counterfactual experimentation. On the EDA-Eval-PyAether benchmark, the system achieved 78.5\% pass@1, substantially outperforming a pure LLM baseline of 23.6\%. However, a critical limitation remained: the agent solved each task independently and did not learn from past execution traces, so its performance on new tasks did not improve over time.

Building on this task-level substrate, we introduce \textbf{self-evolution}: an offline system that distills execution traces into reusable knowledge and re-injects it into future tasks, with reliability gates that prevent unverified experience from misleading the agent. To provide reliable signals for self-evolution, we introduce a precision-optimized static checker for deterministic API misuse. It raises no error flags on the 158 benchmark ground-truth solutions, a sanity check limited to this reference-code set. Applying self-evolution to EDA scripting introduces two challenges. First, feedback is very \textbf{sparse}: the agent receives only a binary pass/fail from the hidden verifier. This is different from methods like SkillOpt~\cite{yang2026skillopt}, which assume dense per-step rewards. Without careful verification, distilled experience can mislead. Second, instance-disjoint evaluation risks cross-split answer copying: a held-out task may copy a training sibling's verbatim solution rather than transferring the underlying method, a problem recently observed in EvolveMem~\cite{evolvemem2026}.

We present \textbf{ZhuLong-SE (Self-Evolution)}, which extends the ZhuLong task-level substrate with enhanced API retrieval and documentation responses, improved sandbox isolation, and precision-optimized static checking, and then adds a self-evolution layer with reliability gating. To handle sparse feedback, we introduce ground-truth anchoring, scope-boundary reflection, disconfirmation-based filtering, and a held-out validation loop that rejects degrading rounds. Beyond post-hoc evaluation gating, we design structural anti-leakage retrieval via names-only injection and deterministic docstring generalization to reduce the risk of cross-split answer copying. These mechanisms are designed to provide methodological signals without injecting full call sequences, while an evolution round is retained only when held-out validation does not degrade beyond the acceptance threshold. On the held-out test set, this design achieves a higher mean pass@1 than the human-crafted-experience baseline.

Our contributions are as follows.

\begin{itemize}[leftmargin=*]
\item \textbf{Self-evolution as a design methodology for EDA automation.} 
We establish self-evolution---distilling execution traces into reusable knowledge and re-injecting them into future tasks---as a systematic methodology for system-level EDA scripting. To make it deployable under EDA's sparse feedback and opaque commercial APIs, we introduce three governance rules: ground-truth anchoring, on-demand pitfall retrieval, and validation-gated rollback.

\item \textbf{A rigorous evaluation protocol against cross-split leakage.} 
We expose cross-split answer copying as a threat to trustworthy self-evolution evaluation, and propose a retrieval-time defense (names-only injection + docstring generalization) that removes full call sequences and task-specific details while preserving methodological transfer---establishing a protocol for future work on self-evolving design agents.

\item \textbf{Counterintuitive design insights from controlled ablation.} 
Controlled ablations on the 158-task EDA benchmark yield three observations: ground-truth anchoring prevents 12 of 24 failed-task traces from producing falsely verified entries; pitfall pre-injection reduces validation pass@1 by 8.0\,pp in our ablation; and names-only retrieval improves over full-sequence injection by 4.7\,pp. On the 40-task held-out test set, self-evolution from the Bare configuration reaches 80.0\% mean pass@1, compared with 77.5\% for the human-crafted-experience baseline, grounding these observations for future system-level design automation.
\end{itemize}

\section{Related Work}
\label{sec:related_work}

\subsection{LLM for EDA Scripting}
\label{sec:rw_eda}

LLMs for chip design have largely targeted RTL generation: ChipNeMo, RTLCoder, AutoChip, and CodeV-R1~\cite{liu2023chipnemo,liu2024rtlcoder,thakur2023autochip,zhu2026qimeng} domain-adapt or fine-tune models for Verilog, relying on human-constructed testbenches, which is orthogonal to operating an EDA tool's API surface. The procedural-scripting strand has grown quickly: ChatEDA, ChaTCL, EDA-Copilot, iScript, FluxEDA, JARVIS, and AutoEDA~\cite{wu2024chateda,rui2025chatcl,xiao2025eda,xu2026iscript,chen2026fluxeda,lu2025autoeda,zhang2025jarvis} span flow planning, Tcl/Q\&A assistance, multi-agent collaboration, single-turn generation, and execution infrastructure. Within this strand, ZhuLong~\cite{liu2026zhulong} is the direct precursor to our work: it combines API retrieval, documentation inspection, sandbox execution, and offline API self-exploration, but solves tasks independently without reusing execution traces across tasks. Another closely related system is OpenROAD Agent~\cite{wu2025openroad}, which adds execution-feedback self-correction on an open physical-design tool, but fine-tunes the model and does not externalize cross-task experience. Our scenario shares its generate-execute-revise shape yet keeps the model frozen and judges pass/fail by real sandbox execution with hidden tests---addressing cross-task reuse through self-evolution rather than fine-tuning. The Agentic EDA survey~\cite{zang2025dawn} situates this line at L3 autonomy. Two gaps recur for a frozen-agent, hidden-test setting: none externalizes cross-task experience into a reliable, leakage-controlled store, and none carries a domain-level static check beyond the tool's own schema---both of which our system addresses.

\subsection{Self-Evolving Agents with Governed Memory}
\label{sec:rw_self_evolving}

A broad family of works equips agents with reusable external memory: skill libraries, cross-task insight distillation, hierarchical memory, and cognitive architectures~\cite{wang2023voyager,zhao2024expel,packer2023memgpt,sumers2023cognitive}, with CodeAct~\cite{wang2024executable} unifying the action space as executable Python. The survey by Zhang et al.~\cite{zhang2024memory} provides a comprehensive taxonomy of memory mechanisms for LLM agents. These works, however, generally assume an open world or a train/test-disjoint split, which sidesteps the same-task regression signal that our fixed-set setting must surface. DemoEvolve~\cite{demoevolve2026} explicitly studies sparse terminal feedback in harness evolution, but addresses it through human demonstrations—whereas our setting assumes no such external guidance.

Treating the skill store as an optimization target, SkillOpt~\cite{yang2026skillopt} edits a single skill document under a held-out gate and observes that ``plausible textual diagnoses can still hurt,'' echoing our concern that unverified experience can mislead. SkillOS~\cite{ouyang2026skillos} learns curation via RL with a four-dimensional judge, and Trace2Skill has been explored in two complementary forms: Ni et al.~\cite{ni2026trace2skill} propose a general framework that distills execution trajectories into reusable skills through parallel trajectory analysis and inductive pattern consolidation, using a verifier primarily for quality filtering of the extracted skills; Du and Pinckney~\cite{du2026trace2skill} present a test-time scaling framework for hardware agents where dense verifier feedback actively drives the oracle-mutator-selector evolution loop. These works optimize on disjoint splits with relatively dense signal, whereas our sparse-pass/fail, fixed-set setting makes which-experience-to-trust more pressing than how to edit a document.

Verifier-guided evolution brings the reward signal closer to our hidden-test setting. ReVeal~\cite{jin2025reveal} self-generates test cases for dense rewards, but self-generated tests can be gamed and do not substitute for a hidden verifier. Du and Pinckney~\cite{du2026trace2skill} share our hidden-verifier, no-answer framing but run online per task with physical isolation and task-wise skills; their finding that cross-task transfer is unreliable for heterogeneous RTL cautions how far logical-error knowledge should travel—a concern we address through governed memory and anti-leakage retrieval. EvoAgentBench~\cite{gao2026evoagentbench}, the most comparable benchmark, formalizes this tension with an ability-supported yet instance-disjoint split, finding that an Anchor Skill oracle helps while no automatic method sustains the gain.

Several works inform the governance machinery underlying our approach: MUSE-Autoskill~\cite{lin2026muse} on refinement and reflection, Externalization~\cite{zhou2026externalization} and Reflexion~\cite{shinn2023reflexion} on memory artifacts and same-task iteration, and Governing Evolving Memory~\cite{lam2026governing} on System-2 verification. LastingBench~\cite{fang2025lastingbench} proposes counterfactual rewriting to safeguard benchmarks against knowledge leakage, underscoring the importance of our anti-leakage retrieval design. The central lesson is that evolving memory requires an external gate grounded in ground truth to remain reliable. Our method addresses this challenge under sparse pass/fail supervision and strict anti-leakage constraints.

\section{Method}
\label{sec:method}

\begin{figure*}[t]
\centering
\includegraphics[width=1.0\textwidth]{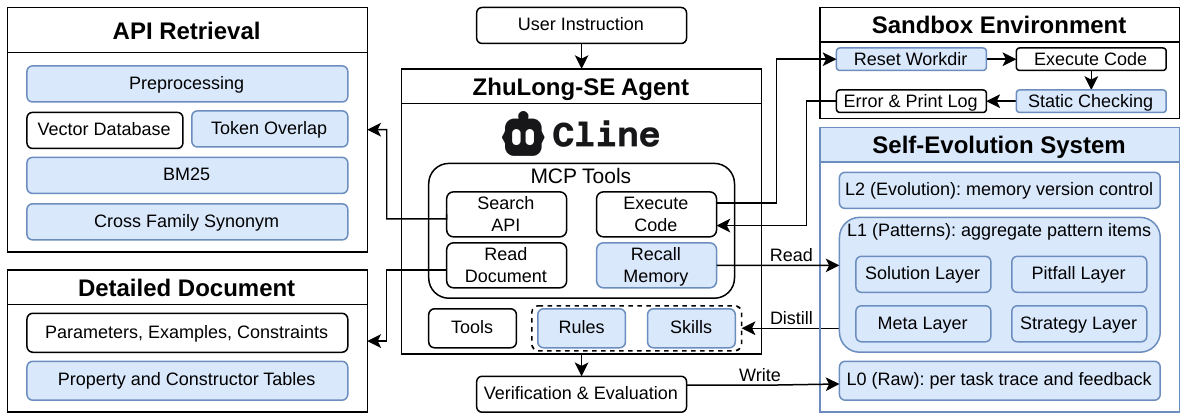}
\caption{Architecture of ZhuLong-SE, adapted from the ZhuLong execution framework~\cite{liu2026zhulong}. Blue-highlighted blocks denote components added or enhanced in this work. Unhighlighted blocks show the shared execution workflow.}
\label{fig:architecture}
\end{figure*}

Building on the origin ZhuLong execution harness~\cite{liu2026zhulong}, ZhuLong-SE retains its retrieval--documentation--execution loop and introduces the blue-highlighted extensions in Figure~\ref{fig:architecture}: enhanced API retrieval and documentation, isolated parallel sandbox management, a precision-optimized static checker, human-crafted rules and skills, and an offline cross-task self-evolution system. We describe the task-level harness extensions in Sections~\ref{sec:method_retrieval}--\ref{sec:method_experience}, followed by self-evolution in Section~\ref{sec:selfev}.

\subsection{API Retrieval}
\label{sec:method_retrieval}

ZhuLong provides a single \texttt{search\_apis} MCP tool that uses semantic retrieval~\cite{liu2026zhulong}. ZhuLong-SE retains its embedding-based semantic path, adds a dedicated keyword-oriented tool (\texttt{search\_apis\_by\_keyword}), and supplements matches missed by the summary with a BM25 pass over the full description text. The keyword path targets queries that the semantic index handles poorly, such as abbreviations and camel-case fragments. The index tokenizes each API name by splitting on camel-case boundaries and stripping family prefixes. The query is bidirectionally expanded for abbreviations (e.g., CV$\leftrightarrow$CellView, crt$\leftrightarrow$create), enabling matches on abbreviated tokens inside API names. Depending on query type, a router expands abbreviations, returns direct-match hints for API-name fragments, or appends semantic results for natural-language queries. Finally, a cross-family synonym map links four functionally equivalent API families (db*, emy*, goa*, ae*), so a hit in one family surfaces equivalents in the others, addressing the failure mode where an agent recalls the correct operation but selects the wrong interface family.

\subsection{Documentation Enrichment}
\label{sec:method_doc}

ZhuLong's \texttt{get\_api\_details} returns parameter types, return values, examples, and inferred constraints~\cite{liu2026zhulong}, but its per-API pages do not expose class instance properties. ZhuLong-SE therefore parses the raw documentation into a property table and appends it to each detailed response, so a queried class exposes its instance attributes alongside its methods. Constructor notes also disambiguate ambiguous calls: for example, \texttt{emyPointArray} accepts a plain list of points, a fact the per-signature page does not state.

\subsection{Sandbox Execution}
\label{sec:method_sandbox}

We run multiple isolated sandbox endpoints, each with an independent working directory, and dispatch requests through a queue pool. A routine clears working directories before each run and resets library files to minimal content, so prior runs do not bias subsequent ones. This isolation ensures that execution feedback reflects the current code under test, not residual state from earlier tasks. The queue pool enables concurrent evaluation without changing ZhuLong's per-task interaction protocol~\cite{liu2026zhulong}.

\subsection{Static Checking}
\label{sec:method_checker}

Because PyAether's SWIG bindings expose all symbols at import time regardless of whether the underlying function exists at runtime, an agent may call a fabricated API or pass a native Python value where a SWIG-wrapped type is required without triggering any runtime error during self-test—yet the submission fails the hidden test that exercises that path. To catch such failures before submission, we introduce a static checker that runs after the agent's self-test and before final submission.

The checker parses submitted code into an abstract syntax tree, extracts API calls, and checks four documented misuse patterns: (i) fabricated APIs, (ii) incorrect keyword arguments, (iii) native Python values passed where explicit SWIG-wrapped construction is required, and (iv) silently swallowed exceptions. Actionable findings are returned to the agent for repair before final submission. The rules are derived from the API documentation and refined on real submissions. As a reference-code sanity check, the checker raises no flags on any of the 158 ground-truth solutions. This result is limited to that set and is not a general zero-false-positive guarantee.

\subsection{Human-Crafted Experience}
\label{sec:method_experience}

ZhuLong-SE additionally injects four skills and one rule file authored from PyAether development experience rather than specific benchmark solutions. The skills cover API search, command-family selection, interpretation of silent or ambiguous sandbox feedback, and explicit SWIG type construction. The rule file requires API-detail checks before use, post-creation verification, and bounded search, providing a stable prior that complements knowledge learned from the execution traces.

\section{Self-Evolution}
\label{sec:selfev}

Unlike ZhuLong's offline API self-exploration, which augments per-API documentation through counterfactual sandboxing~\cite{liu2026zhulong}, ZhuLong-SE adds a cross-task offline self-evolution system (the blue-highlighted block at right in Figure~\ref{fig:architecture}). It distills task execution traces into reusable knowledge and re-injects it into later tasks. Knowledge is organized in three tiers, following the episodic-semantic-procedural memory distinction~\cite{sumers2023cognitive}: \emph{L0} holds raw traces, \emph{L1} stores aggregated patterns, and \emph{L2} archives distilled skill versions. Within L1, patterns are partitioned into four layers: \emph{solution} (verified API routes), \emph{pitfall} (negative warnings), \emph{meta} (effective query patterns), and \emph{strategy} (decision records). The write side runs offline after each training batch, and the read side injects knowledge at task formatting and on demand when the agent hits an error. The write, promotion, and round-acceptance stages apply reliability gates, as unvalidated experience can mislead~\cite{zhao2024expel,yang2026skillopt}. Distillation and state updates run offline. During inference, the accepted memory is queried locally, without additional evolution or distillation.

\subsection{Distillation and Reflection}
\label{sec:se_distill}

The write side routes each task to the appropriate reflection path. We override the agent's self-reported success with the evaluator's binary verdict before any reflection begins, following Reflexion's observation that reliable feedback is necessary for reflection to help rather than harm~\cite{shinn2023reflexion}. Tasks that still error on their final attempt take the \emph{error path}; tasks whose self-test passed but the evaluator failed take the \emph{contradiction path} instead. This distinction is critical because a self-test-passed task's error log contains only stale errors; mining pitfalls from these would record already-solved bugs rather than genuine gaps in the agent's knowledge. Each L1 entry carries a reliability tier: \emph{verified}, \emph{unverified}, or \emph{pitfall}.

\subsection{Knowledge Lifecycle with Reliability Gating}
\label{sec:se_govern}

Extracted L1 patterns undergo aggregation, gating, aging, and promotion to rules and skills. A verified entry is disconfirmed only when a failed task is an API misuse precisely attributed to that entry; at a count of two or more, the entry is filtered out. Unverified traces cannot downgrade verified entries, and entries not re-confirmed for five rounds drop to unverified. Promotion is guarded by a four-dimensional judge (abstraction, reusability, actionability, faithfulness) following SkillOS~\cite{ouyang2026skillos}, and a MUSE-style create-evaluate-refine loop~\cite{lin2026muse} for low-scoring candidates. We add a zero-exception generality gate that rejects rules whose always/never statements do not hold across the entire API surface, ensuring that distilled rules remain broadly applicable rather than overfitting to specific task patterns.

\subsection{Anti-Leakage Retrieval}
\label{sec:se_read}

Under an instance-disjoint split, naive retrieval risks cross-split answer copying. We prevent this through three mechanisms. \textbf{Names-only pre-injection} reduces a solution entry to API names only, dropping call order and parameters. \textbf{Deterministic docstring generalization} strips task-identifying specifics from both distillation and query sides. \textbf{Problem-side embedding} matches on problem similarity while returning solution metadata only at read time. Pitfalls are not pre-injected; they enter through on-demand runtime queries keyed to the agent's actual error.

\subsection{Held-Out Gated Evolution}
\label{sec:se_loop}

After each training round, the agent generates on validation tasks (which never carry distillation). The round's full evolved state is accepted only if the validation pass rate does not regress by more than 5~pp against the previous round. Otherwise, it is rolled back. This extends SkillOpt's held-out criterion~\cite{yang2026skillopt} from a single document to the entire evolved state.

\section{Experiments}
\label{sec:experiments}

We evaluate ZhuLong-SE on EDA-Eval-PyAether, the 158-task real-world PyAether scripting benchmark introduced by ZhuLong~\cite{liu2026zhulong}\footnote{We obtained the benchmark by contacting its authors.}. Each task is a natural-language docstring plus a function signature, and a submission passes only if real sandbox execution satisfies a hidden test (the black-box verifier). The primary model is DeepSeek-V4-Pro-Preview. The human-crafted experience consists of four skills and one rule authored by a senior EDA engineer with extensive experience in PyAether development.

The experiments have two parts. First, on all 158 tasks, we ask whether the enhanced ZhuLong-SE harness outperforms the original ZhuLong across matched LLM backbones and which measured harness components contribute to the gain. Second, after removing human-crafted experience, we ask whether governed self-evolution can match or surpass the human-crafted-experience baseline.

\subsection{Harness Evaluation}
\label{sec:exp_retrieval}

We first evaluate the enhanced task-level harness, with self-evolution disabled throughout this subsection. The complete ZhuLong-SE configuration enables all harness enhancements described in Section~\ref{sec:method}, including full retrieval, documentation enrichment, sandbox isolation, the static checker, and human-crafted experience. We first compare this configuration with the published ZhuLong baseline and then decompose the measured components under the primary backbone. No self-evolved knowledge is injected into any result reported in this subsection. For each comparison, the backbone version and settings other than the harness components under study are aligned with ZhuLong~\cite{liu2026zhulong}. Table~\ref{tab:cross_method} uses the reported ZhuLong results where available.

Table~\ref{tab:cross_method} compares ZhuLong-SE with ZhuLong on the same backbone versions. Across all the five comparisons, ZhuLong-SE improves by 11.4~pp on DeepSeek-V4-Pro-Preview, 16.4~pp on GLM-5.2, 5.7~pp on Kimi-K2.6, 6.9~pp on DeepSeek-V4-Flash-Preview, and 15.8~pp on Doubao-Seed-2.0-pro. Because self-evolution is disabled, these gains reflect the combined task-level enhancements in Section~\ref{sec:method}. The comparisons indicate transfer across the tested model families, although absolute performance still depends on backbone capability.

\begin{table}[ht]
\caption{Harness-only Pass@1 comparison with ZhuLong on aligned backbone versions (158 tasks, one ZhuLong-SE run per model, no self-evolution). }
\label{tab:cross_method}
\setlength{\tabcolsep}{2pt}
\centering
\small
\begin{tabular}{lccr}
\toprule
Model & ZhuLong & ZhuLong-SE & $\Delta$ \\
\midrule
DeepSeek-V4-Pro-Preview   & 81.0\% & 92.4\% & +11.4~pp \\
GLM-5.2                   & 74.7\% & 91.1\% & +16.4~pp \\
Kimi-K2.6                 & 83.5\% & 89.2\% & +5.7~pp \\
DeepSeek-V4-Flash-Preview & 78.5\% & 85.4\% & +6.9~pp \\
Doubao-Seed-2.0-pro       & 55.7\% & 71.5\% & +15.8~pp \\
\bottomrule
\end{tabular}
\end{table}

Each harness configuration is evaluated once because a full 158-task run is costly, and the broad task coverage reduces sensitivity to any single task outcome. Repeated runs are reserved for the smaller validation and challenge-test evaluations.


We next perform an ablation study on ZhuLong-SE. Because ZhuLong-SE is inherited from ZhuLong, we use progressive rather than leave-one-out ablation to attribute that advantage to the newly added components. Checker \emph{ON} enables the static checker, whereas \emph{OFF} bypasses it. \emph{Semantic-Only} retains \texttt{search\_apis} but disables the keyword tool and BM25 supplement, whereas \emph{Full} enables both tools. \emph{Bare} uses no human-crafted knowledge, whereas \emph{Human} adds four skills and one rule derived from human PyAether development experience rather than the benchmark tasks. DeepSeek-V4-Pro-Preview serves as the fixed inference backbone. Table~\ref{tab:ablation} shows that successive additions raise Pass@1 from 81.0\% to 87.3\% with the checker (+6.3~pp), to 88.6\% with Full retrieval (+1.3~pp), and to 92.4\% with Human experience (+3.8~pp). Each component therefore contributes, with the checker providing the largest stepwise gain. Documentation enrichment and sandbox isolation remain fixed and are evaluated only in the overall harness comparison.

\begin{table}[ht]
\caption{Progressive harness ablation on the 158-task benchmark (single run, DeepSeek-V4-Pro-Preview).}
\label{tab:ablation}
\centering
\small
\begin{tabular}{lccr}
\toprule
Experience & Retrieval & Checker & Pass@1 \\
\midrule
Bare  & Semantic-Only & OFF & 81.0\% \\
Bare  & Semantic-Only & ON & 87.3\% \\
Bare  & Full          & ON & 88.6\% \\
Human & Full          & ON & 92.4\% \\
\bottomrule
\end{tabular}
\end{table}


Task~132 illustrates why the keyword path complements semantic retrieval. The task connects two orthogonal routing paths by placing a via at their intersection and requires constructing several SWIG-wrapped types. With Semantic-Only retrieval, the agent issued natural-language queries but did not retrieve the required type definitions, eventually passing a native tuple where a wrapped object was required. With Full retrieval, the agent queried \texttt{ViaDef}, \texttt{emyPoint}, and \texttt{emyString} through the keyword path, retrieved their construction signatures, and completed the task successfully. This case reflects the intended role of camel-case splitting and exact-fragment matching: they make abbreviation-dense API and type names directly addressable when natural-language similarity is insufficient.

\begin{figure}[ht]
\centering
\includegraphics[width=\columnwidth]{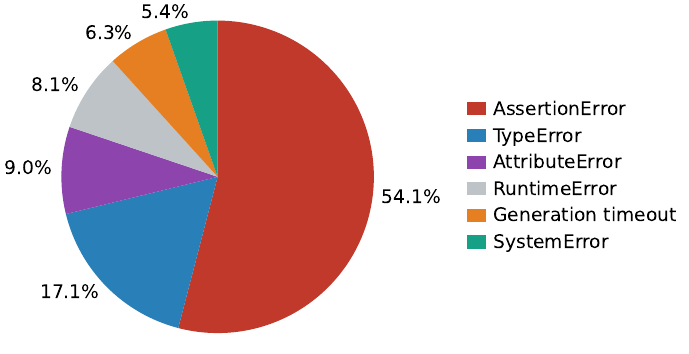}
\caption{Residual failure types of ZhuLong-SE across five LLM backbones on the 158-task benchmark (111 failures aggregated).}
\label{fig:failure_types}
\end{figure}

Finally, we characterize the residual failures of the enhanced harness. Figure~\ref{fig:failure_types} aggregates the 111 failed tasks across the five ZhuLong-SE backbone runs in Table~\ref{tab:cross_method}. \texttt{AssertionError} accounts for 54.1\% of these failures: the generated program reaches the hidden verifier without an uncaught API or type error, but its result violates a tested property. Such semantic or result-level errors lie outside the static checker's declared scope, which targets deterministic API misuse rather than hidden functional correctness. This analysis therefore characterizes the remaining failure frontier rather than directly measuring checker recall. Since the task-level harness still solves each task independently, these residual errors motivate the next subsection's evaluation of whether validated knowledge distilled from prior traces can improve later tasks.

\subsection{Self-Evolution}
\label{sec:exp_selfev}

To evaluate self-evolution without ceiling effects—the Bare agent already passes 87\% of the 158 tasks—and to support held-out gating, we adopt an instance-disjoint \emph{challenge split} (train~98 / val~20 / test~40), in the spirit of EvoAgentBench's ability-supported yet instance-disjoint split~\cite{gao2026evoagentbench}. The split is stratified so that the test set concentrates more discriminative tasks, identified by testbench code length and the number of unique APIs used. Because the validation and test sets are small, every self-evolution evaluation is the mean of \textbf{five} independent runs with its standard deviation.

Table~\ref{tab:sevolution} reports test Pass@1 across rounds, against the Bare baseline (no evolved knowledge) and the Human baseline. Self-evolution clears the Bare baseline by +14.0~pp (66.0\%$\to$80.0\%) and surpasses the Human baseline (80.0\% vs.\ 77.5\%) at Round~2, after which Round~3 regresses to 75.5\%. The validation trend is consistent: validation peaks at Round~2 (88.0\%) and drops at Round~3 (80.0\%), so the gate accepts Rounds~1 and~2 but rejects Round~3, rolling back its 52 new entries and freezing the knowledge base at Round~2 (334 entries, 20 skills). Crucially, the test trend agrees with validation—Round~3 regresses on both (75.5\% vs.\ 80.0\%)—confirming the gate's rejection reflects genuine overfitting, not a validation-specific artifact. Round~2 also has the smallest observed run-to-run spread ($\pm$3.1), below the Human baseline's $\pm$5.3.

\begin{table}[ht]
\caption{Self-evolution on the challenge test set (40 tasks, 5 runs each, DeepSeek-V4-Pro-Preview). ``Patterns'' is the L1 entry count, ``skills'' are auto-distilled. The held-out gate accepted Rounds~1 and~2, and rejected Round~3.}
\label{tab:sevolution}
\setlength{\tabcolsep}{2.5pt}
\centering
\small
\begin{tabular}{lcccc}
\toprule
Knowledge & Patterns & Skills & Pass@1 (test) & Pass@1 (val) \\
\midrule
Bare          & 0   & 0  & 66.0\%$\pm$7.2  & --- \\
Round~1 & 256 & 16 & 78.0\%$\pm$9.9  & 81.0\%$\pm$8.2 \\
Round~2 & 334 & 20 & \textbf{80.0\%}$\pm$3.1  & 88.0\%$\pm$2.7 \\
Round~3 & 386 & 25 & 75.5\%$\pm$6.7  & 80.0\%$\pm$3.5 \\
\midrule
Human Baseline & --- & 4  & 77.5\%$\pm$5.3  & --- \\
\bottomrule
\end{tabular}
\end{table}

To examine the write-side gates under a fixed input, we freeze the 24 distinct failed-task traces accumulated across post-split batches and re-run the offline reflection chain with each gate toggled. This avoids the cost and sparse signal of full end-to-end ablation (13 hours per round), following Trace2Skill's shared-trace-pool design~\cite{ni2026trace2skill}. Test failure traces serve only as offline-reflection inputs, and knowledge produced from them is never injected into test evaluation. Table~\ref{tab:se_write_ablation} reports the resulting L1 entry counts. Because reflection is not fully deterministic, we interpret only gate-shaped changes rather than one-entry differences.

\begin{table}[ht]
\caption{Write-side gate ablation on the frozen 24-task failure trace pool.
``ON'' and ``OFF'' report L1 entry counts.}
\label{tab:se_write_ablation}
\centering
\begin{tabular}{lcc}
\toprule
Gate & ON & OFF \\
\midrule
Ground-Truth Anchoring   & 37 & 48 \\
Contradiction Path       & 37 & 54 \\
Routing                  & 36 & 37 \\
Scope Boundary           & 36 & 36$^{*}$ \\
Disconfirmation Filter   & 37 & 37$^{\dagger}$ \\
\bottomrule
\end{tabular}

\vspace{2pt}
{\footnotesize\raggedright
$^{*}$ Affects entry content rather than count.
$^{\dagger}$ Inactive because the threshold requires disconfirmation
across at least two batches.\par}
\end{table}

Ground-truth anchoring has the clearest observed effect: without it, 12 of 24 tasks yield at least one falsely verified entry because the trace's self-reported success is trusted. The net count rises by 11 rather than 12 because one such task produces no entry under correct anchoring. Disabling the contradiction path produces 17 additional entries by mining stale, already-repaired errors, while disabling routing sends failed tasks to the success branch and produces no pitfalls. These entry-count results characterize how the three gates alter reflection output. They do not constitute an end-to-end quality comparison.

The remaining two write-side gates cannot be evaluated by this design. Scope-boundary reflection changes the preconditions and constraints recorded within an entry rather than the entry count, so the reported metric cannot identify its effect. The disconfirmation filter removes an entry only after disconfirmation in at least two batches; no entry can reach that threshold in this single-round replay. Their value therefore remains unverified here and requires content-level assessment and multi-round accumulation, respectively.

We also ablate each of the four read-side mechanisms from Section~\ref{sec:se_read} (names-only pre-injection, docstring generalization, problem-side embedding, and the decision not to pre-inject pitfalls) by freezing the Round~2 knowledge and varying the injection configuration, then re-running the agent on validation (baseline 88.0\%$\pm$2.7). Table~\ref{tab:se_read_ablation} isolates each. Pre-injecting pitfalls is the most harmful ($-$8.0~pp): a task's description does not predict which API it will misuse, so a recalled pitfall is usually off-target, confirming the choice to route pitfalls through the runtime query instead. Docstring generalization ($-$6.3~pp when disabled) and names-only injection ($-$4.7~pp when full-sequence injection is used) each contribute—the former because disabling it breaks the query-embedding symmetry, the latter because a full call sequence over-specifies and misleads. The embedding strategy (embedding the problem summary vs.\ the solution text) shows no measurable effect here, which we attribute to names-only injection: under this reduction, the retrieved content is the same regardless of embedding, so the embedding choice is dominated by the names-only reduction. The problem-side embedding is therefore justified by its anti-leakage role rather than by a measurable Pass@1 gain in this configuration.

\begin{table}[ht]
\caption{Read-side injection ablation, frozen Round~2 knowledge, validation (20 tasks, 5 runs). Each row varies one mechanism away from the full system (baseline 88.0\%$\pm$2.7).}
\label{tab:se_read_ablation}
\centering
\small
\begin{tabular}{lcc}
\toprule
Configuration & Pass@1 (val) & $\Delta$ \\
\midrule
Full System (Baseline)        & 88.0\%$\pm$2.7 & --- \\
Pitfall Pre-Injection         & 80.0\%$\pm$5.0 & $-8.0$~pp \\
Docstring Generalization Off  & 81.7\%$\pm$5.8 & $-6.3$~pp \\
Full-Sequence Injection       & 83.3\%$\pm$5.8 & $-4.7$~pp \\
Embed Solution (Not Problem)  & 88.0\%$\pm$2.7 & $0$~pp \\
\bottomrule
\end{tabular}
\end{table}

\section{Conclusion and Future Work}

We presented \textbf{ZhuLong-SE (Self-Evolution)}, which augments ZhuLong~\cite{liu2026zhulong} with task-level harness enhancements and governed cross-task evolution while keeping the backbone frozen. On the 158-task EDA-Eval-PyAether benchmark, the full task-level harness achieves 92.4\% pass@1 in a single full run and outperforms ZhuLong on all five backbone comparisons. On the 40-task instance-disjoint challenge test, self-evolution from the Bare configuration achieves 80.0\% mean pass@1, compared with 77.5\% for the human-crafted-experience baseline. The frozen-trace ablation identifies the effects of ground-truth anchoring, contradiction handling, and routing on reflection output, while leaving scope-boundary reflection and multi-batch disconfirmation unverified under its entry-count, single-round design. Future work will extend the evaluation to additional EDA tool flows and languages, including SKILL-based automation, and to larger task sets, while investigating adaptive gating thresholds that adjust to task difficulty.

\balance
\bibliographystyle{IEEEtran}
\bibliography{reference}

\end{document}